\documentclass[sigplan,screen]{acmart}

\usepackage{adjustbox}
\usepackage{amsthm}
\usepackage{algorithm}
\usepackage[noend]{algpseudocode}
\usepackage{colortbl}
\usepackage{graphicx}
\usepackage{mathtools}
\usepackage{pifont}
\usepackage{threeparttable}
\usepackage{multirow}
\usepackage{multicol}
\usepackage{marvosym}

\newcommand{\Neuralink}{\textsc{Neuralink}}
\makeatletter
\NewDocumentCommand{\LeftComment}{s m}{
  \Statex \IfBooleanF{#1}{\hspace*{\ALG@thistlm}}\(\triangleright\) #2}
\makeatother

\copyrightyear{2025}
\acmYear{2025}
\setcopyright{cc}
\setcctype{by-nc-sa}
\acmConference[ASPLOS '25]{Proceedings of the 30th ACM International Conference on Architectural Support for Programming Languages and Operating Systems, Volume 3}{March 30-April 3, 2025}{Rotterdam, Netherlands}
\acmBooktitle{Proceedings of the 30th ACM International Conference on Architectural Support for Programming Languages and Operating Systems, Volume 3 (ASPLOS '25), March 30-April 3, 2025, Rotterdam, Netherlands}
\acmDOI{10.1145/3676642.3736114}
\acmISBN{979-8-4007-1080-3/2025/03}

\begin{document}

\title{\Neuralink{}: Fast LLM Inference on Smartphones with Neuron Co-Activation Linking}



\author{Tuowei Wang}
\authornote{Both authors contributed equally to this research.}
\affiliation{%
  \institution{Tsinghua University}
  \city{Beijing}
  \country{China}
}

\author{Ruwen Fan}
\authornotemark[1]
\affiliation{%
  \institution{Tsinghua University}
  \city{Beijing}
  \country{China}
}

\author{Minxing Huang}
\affiliation{%
  \institution{Tianjin University}
  \city{Tianjin}
  \country{China}
}

\author{Zixu Hao}
\affiliation{%
  \institution{Tsinghua University}
  \city{Beijing}
  \country{China}
}

\author{Kun Li}
\affiliation{%
  \institution{Microsoft Research}
  \city{Beijing}
  \country{China}
}

\author{Ting Cao}
\affiliation{%
  \institution{Microsoft Research}
  \city{Beijing}
  \country{China}
}

\author{Youyou Lu}
\affiliation{%
  \institution{Tsinghua University}
  \city{Beijing}
  \country{China}
}

\author{Yaoxue Zhang}
\affiliation{%
  \institution{Tsinghua University}
  \city{Beijing}
  \country{China}
}

\author{Ju Ren}
\authornote{Corresponding author (\href{mailto:renju@tsinghua.edu.cn}{renju@tsinghua.edu.cn}).}
\affiliation{%
  \institution{Tsinghua University}
  \city{Beijing}
  \country{China}
}

\renewcommand{\authors}{Tuowei Wang, Ruwen Fan, Minxing Huang, Zixu Hao, Kun Li, Ting Cao, Youyou Lu, Yaoxue Zhang, Ju Ren}
\renewcommand{\shortauthors}{Tuowei Wang et al.}

\begin{abstract}
Large Language Models (LLMs) have achieved remarkable success across various domains, yet deploying them on mobile devices remains an arduous challenge due to their extensive computational and memory demands. While lightweight LLMs have been developed to fit mobile environments, they suffer from degraded model accuracy. In contrast, sparsity-based techniques minimize DRAM usage by selectively transferring only relevant neurons to DRAM while retaining the full model in external storage, such as flash. However, such approaches are critically limited by numerous I/O operations, particularly on smartphones with severe IOPS constraints.

In this paper, we propose \Neuralink{}, a novel approach that accelerates LLM inference on smartphones by optimizing neuron placement in flash memory. \Neuralink{} leverages the concept of \textit{Neuron Co-Activation}, where neurons frequently activated together are linked to facilitate continuous read access and optimize I/O efficiency. Our approach incorporates a two-stage solution: an offline stage that reorganizes neuron placement based on co-activation patterns, and an online stage that employs tailored data access and caching strategies to align well with hardware characteristics. Evaluations conducted on a variety of smartphones and LLMs demonstrate that \Neuralink{} achieves on average $1.49\times$ improvements in end-to-end latency compared to the state-of-the-art. As the first solution to optimize storage placement under sparsity, \Neuralink{} explores a new optimization space at the intersection of sparsity-driven algorithm and storage-level system co-design for LLM inference.
\end{abstract}

\ccsdesc[500]{Computer systems organization~Heterogeneous (hybrid) systems}
\ccsdesc[500]{Information systems~Storage management}

\keywords{Mobile Computing, Large Language Model, Model Sparsity, Parameter Storage}

\maketitle 

\section{Introduction} 
Large Language Models (LLMs) have demonstrated exceptional performance across a wide range of applications~\cite{chatgpt, gpt4, llm-application-medical,llm-application-education,llm-application-finance,llm-application-engineer}. Comprising millions or even billions of parameters~\cite{bert,opt,gpt3,llama2,palm,mistral,gemini}, these models necessitate substantial computational and memory resources, routinely available only in cutting-edge data centers. Nonetheless, there is a growing demand for deploying LLMs on resource-constrained devices, such as smartphones~\cite{llm-mobile-1,llm-mobile-2,llm-mobile-3,llm-mobile-4,llm-mobile-5,llm-mobile-6}. On one hand, stringent privacy regulations mandate local data processing to protect user information. On the other hand, on-device inference reduces the latency incurred by cloud-based computations, enabling real-time responses.

Given the limited DRAM capacity of devices, LLMs deployed on smartphones are mostly restricted to models tailored for mobile deployment~\cite{phi3,mobillama,minicpm,gemini}. Although these models are designed to be lightweight, the reduction in parameters inevitably leads to a compromise in their capabilities~\cite{scaling-law}. In response, several recent studies~\cite{powerinfer-2,llm-flash,deja-vu,powerinfer,sparsity-mobile-survey-1,sparsity-mobile-survey-2} propose leveraging the intrinsic sparsity within LLMs, particularly \textit{Activation Sparsity}. Similar to how the human brain does not activate all its neurons at once, these approaches activate only a subset of model parameters while still producing the same output as their dense counterparts for a given input. Therefore, larger and more powerful LLMs can be stored in \textbf{flash memory}, with only the activated parameters being transferred to \textbf{DRAM} for computation. Considering the much larger capacity of flash memory, the DRAM limitations of smartphones are effectively offset.

\begin{table}[b]
    \small
    \centering
    \caption{Breakdown of inference latency per token when offloading FFN blocks to flash memory on OnePlus Ace2.}
    \label{tab:intro-io-bottleneck}
    \begin{adjustbox}{width=1.0\linewidth,center}
    \setlength{\tabcolsep}{4pt}
    \begin{tabular}{lrrrr}
        \toprule
        \textbf{Model} & \textbf{Compute}   & \textbf{I/O}       & \textbf{Total}     & \textbf{I/O Ratio} \\ \midrule
        OPT-350M       & 82 ms              & 776 ms             & 858 ms             & 90.4\%             \\
        OPT-1.3B       & 202 ms             & 988 ms             & 1,190 ms           & 83.0\%             \\
        OPT-6.7B       & 804 ms             & 2,224 ms           & 3,028 ms           & 73.4\%             \\
        Llama-2-7B     & 609 ms             & 10,388 ms          & 10,997 ms          & 94.5\%             \\
        Mistral-7B     & 540 ms             & 12,220 ms          & 12,760 ms          & 95.8\%             \\
        MobiLlama-1B   & 230 ms             & 1,909 ms           & 2,139 ms           & 89.2\%             \\
        Phi-2-2.7B     & 461 ms             & 1,976 ms           & 2,437 ms           & 81.1\%             \\
        \bottomrule
    \end{tabular}
    \end{adjustbox}
\end{table}

\begin{figure}[t]
    \centering
    \includegraphics[width=1.0\linewidth]{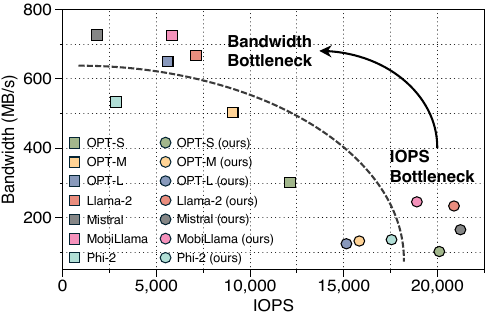}
    \caption{The bandwidth and IOPS during inference across various LLMs on OnePlus Ace2. \Neuralink{} shifts the I/O bottleneck from IOPS (lower right) to bandwidth (upper left).}
    \label{fig:intro-core-result}
\end{figure}

Unfortunately, I/O overheads severely impede the efficiency of the activation sparsity paradigm. Due to the disjoint sets of activated parameters across different inference requests, frequent I/O operations are necessitated to swap model parameters between DRAM and flash memory. As detailed in Table~\ref{tab:intro-io-bottleneck}, I/O operations account for 73.4\%-95.8\% of the total inference latency when offloading only the Feed Forward Network (FFN) of models to flash memory. Especially, rather than being bound on bandwidth, the I/O efficiency on smartphones is primarily bottlenecked by \textbf{Input/Output Operations Per Second (IOPS)}~\cite{iops}, as depicted in Figure~\ref{fig:intro-core-result}. On one hand, flash memory on smartphones generally features a shallow I/O command queue. More critically, the scattered activation of parameters induces numerous small-grained read accesses, further intensifying the constraint.

In this paper, we present a key insight that neurons in LLMs prevalently exhibit strong correlations in their activation patterns, which can be strategically leveraged to accelerate I/O operations. Specifically, when processing real-world datasets, the activation of an individual neuron is consistently accompanied by the activation of a stable set of other neurons, a phenomenon we term \textit{Neuron Co-Activation}. By taking advantage of continuous reads, which enable the retrieval of larger data blocks with a single request, we can \textit{co-locate frequently co-activated neurons in contiguous memory addresses in flash memory}, reducing IOPS and thereby improving the overall inference efficiency on smartphones.

However, this is not a low-hanging fruit. Prior works~\cite{deja-vu,powerinfer} primarily emphasize computation efficiency under activation sparsity, inadvertently exacerbating I/O bottlenecks. Fewer studies~\cite{powerinfer-2,llm-flash} develop advanced caching strategies to reduce I/O volume, leaving I/O bandwidth still held back by IOPS constraints. As an orthogonal approach, directly improving bandwidth utilization requires an effective alignment between both neuron co-activation patterns and storage hardware characteristics. Our comprehensive analysis identifies three critical challenges that must be tackled:

\noindent(1) \textbf{Extensive Search Space.} The immense number of neurons in LLM results in an exponentially large space of possible placement combinations. Determining the optimized placement that maximizes global benefits is highly challenging and infeasible through brute-force enumeration alone.

\noindent(2) \textbf{Inherent Activation Dynamics.} The activation patterns of parameters inherently exhibit dynamics across varying inputs. Although optimized placement strategies can spatially co-locate activated neurons, accesses to these neurons remain impeded by discontinuities induced by dynamics.

\noindent(3) \textbf{Misaligned Cache Strategy}. Caching frequently activated neurons in memory is essential for reducing I/O workload. Existing cache strategies typically treat neurons individually, which can lead to fragmentation in their placement in flash memory, potentially disrupting continuous access.

To address these emerging challenges, we propose \Neuralink{}, a novel approach to accelerating LLM inference on smartphones with neuron co-activation linking. Specifically, \Neuralink{} employs a two-stage solution that incorporates hierarchical optimizations performed both offline and online.

\noindent(1) \textbf{In the Offline Phase}, \Neuralink{} clusters neurons exhibiting high co-activation correlation and reorganizes their placement in flash memory. To address \textbf{Challenge (1)}, we abstract the problem into a complete graph, reformulating it as the discovery of the globally optimal Hamiltonian Path. By leveraging graph-theoretic techniques, we propose a greedy algorithm that efficiently searches for optimized placement based on observed neuron co-activation patterns.

\noindent(2) \textbf{In the Online Phase}, \Neuralink{} performs fine-grained refinements on optimized neuron placement, further enhancing access continuity. To tackle \textbf{Challenge (2)}, we devise an IOPS-friendly access collapse technique. By strategically incorporating additional neurons between two separate neuron links, we improve read access continuity with negligible overhead. In response to \textbf{Challenge (3)}, we introduce a linking-aligned in-memory caching policy. Rather than individually caching the hottest neurons, we account for their interlinking relationships, ensuring continuous access.

We evaluate \Neuralink{} on three smartphones with distinct hardware configurations, benchmarking a diverse range of LLMs varying in structures and scales. The results demonstrate that \Neuralink{} significantly enhances I/O bandwidth, achieving improvements of $1.80\times$ on average. Moreover, this boost in bandwidth yields substantial reductions in end-to-end latency during inference, delivering average speedups of $1.49\times$ when compared to state-of-the-art solutions.

To the best of our knowledge, \Neuralink{} is the first to accelerate LLM inference on smartphones by enhancing I/O bandwidth through optimized neuron placement in flash memory. Our contributions can be summarized as follows:
\begin{itemize}
    \item We identify the primary bottleneck in LLM inference on smartphones as IOPS, attributing it to the inherent misalignment between scattered activation patterns and storage hardware characteristics.
    \item We notably exploit neuron co-activation to mitigate the IOPS bottleneck, pioneering the optimization of neuron placement in flash memory on smartphones.
    \item We conduct extensive evaluations on various representative LLMs and hardware, achieving substantial improvements over state-of-the-art solutions.
\end{itemize}

\section{Background and Motivation}
\subsection{Activation Sparsity in LLM Inference}
Numerous studies~\cite{lazy-neuron,inference-activation-sparsity,slm-activation-sparsity,deja-vu,long-exposure} have shown that LLMs exhibit considerable \textit{Activation Sparsity}, allowing a substantial portion of parameters to be deactivated without impacting the final outputs. As only a subset of parameters is involved in the computation, this characteristic greatly reduces resource consumption. Importantly, since no parameters are pruned, the full capacity of the LLMs remains intact.

\begin{figure}[t]
    \centering
    \includegraphics[width=1.0\linewidth]{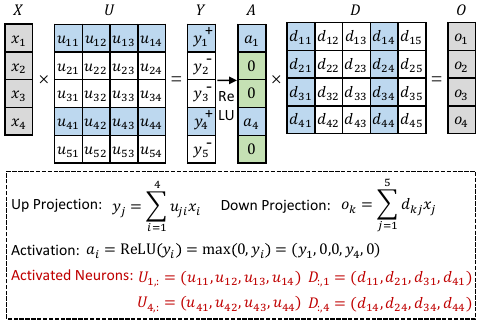}
    \caption{Activation sparsity introduced by ReLU. Each element in the intermediate activations $A$ with a zero value (colored in green) deactivates two neurons (uncolored): the corresponding row in up-projection matrix $U$ and the column in down-projection matrix $D$ within the FFN block.}
    \label{fig:background-activation-sparsity}
\end{figure}

Figure~\ref{fig:background-activation-sparsity} illustrates an example of activation sparsity within the FFN block introduced by ReLU function~\cite{relu}. For a given input $X$, the ReLU function sets the negative values of intermediate activations $Y$ to zero, rendering the corresponding neurons in both up projection weights $U$ and down projection weights $D$ unnecessary. Therefore, these neurons can be considered non-activated and effectively excluded from computation without impacting the model outputs. In addition to ReLU-family LLMs, several studies have highlighted that LLMs using other activation functions can also benefit from activation sparsity through techniques such as ReLU substitution~\cite{relu-strikes-back,pro-sparse,turbo-sparse}, moefication~\cite{moefication,breaking-relu-barrier}, top-K sparsity~\cite{q-sparse,lazy-neuron}, threshold~\cite{chess,cats}, and attribution score~\cite{slm-activation-sparsity,relu-2-wins}.

\begin{figure}[t]
    \centering
    \includegraphics[width=1.0\linewidth]{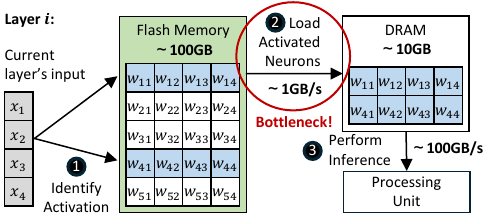}
    \caption{A three-step procedure for LLM inference on smartphones leveraging activation sparsity: \ding{182} Identify the activated neurons for a given input using predictors~\cite{deja-vu,powerinfer} or sparsity-aware metrics~\cite{q-sparse,slm-activation-sparsity}. \ding{183} With the full parameters stored in flash memory, load only the activated neurons into DRAM. \ding{184} Perform inference using the activated neurons.}
    \label{fig:background-paradigm}
\end{figure}

Particularly, activation sparsity offers a promising solution to overcome DRAM limitations on smartphones~\cite{llm-flash,powerinfer-2}. As illustrated in Figure~\ref{fig:background-paradigm}, the full model parameters can be stored in much larger flash memory, while only the activated neurons are loaded into DRAM. Therefore, larger and more powerful LLMs can be executed under limited DRAM capacity, which is crucial for LLM deployment on smartphones. On one hand, smartphones typically provide limited DRAM capacity, ranging from 10GB to 20GB. On the other hand, a substantial portion of DRAM is allocated to the operating system and other active applications, leaving even less available for any single application. Although activation sparsity alleviates this issue by reducing DRAM demands, it introduces a new challenge: the I/O overhead between flash memory and DRAM becomes the primary bottleneck.

\begin{figure}[t]
    \centering
    \includegraphics[width=1.0\linewidth]{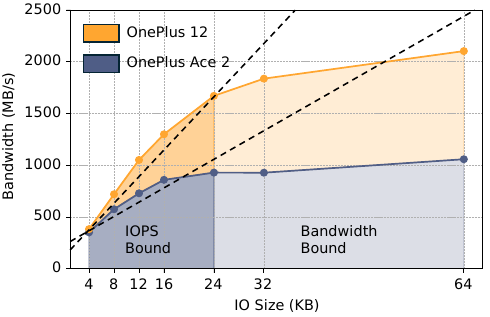}
    \caption{UFS Bandwidth at varying continuous I/O sizes on smartphones. The near-linear relationship indicates that the bottleneck lies in IOPS, rather than the bandwidth capacity.}
    \label{fig:background-iops}
\end{figure}

\subsection{Universal Flash Storage on Smartphones}
Mobile devices, such as smartphones, predominantly utilize Universal Flash Storage (UFS)~\cite{ufs} as the storage protocol. Leveraging NAND flash, UFS provides significantly larger storage capacity than the space available in DRAM, with scalability reaching terabyte (TB) levels. Furthermore, the introduction of the command queue in UFS markedly improves the efficiency of data transfer between flash memory and DRAM. In the latest version (UFS 4.0), the sustained read speed per lane can reach up to 2.9 GB/s. This combination of extensive storage capacity and high read speed serves as a foundation for the execution of LLMs on mobile devices.

However, unlike server-side storage (such as NVMe), UFS typically features a shallow command queue, supporting only 32 entries. This limitation restricts the IOPS for flash reads and prevents full utilization of the available bandwidth. As shown in Figure~\ref{fig:background-iops}, the read bandwidth increases with the continuous I/O sizes, as multiple continuous reads can be issued by a single read operation. Notably, when the continuous I/O size is below 24KB, bandwidth scales almost linearly with I/O size, indicating that these reads are primarily IOPS-bound. Therefore, the key to fully exploiting UFS bandwidth lies in maximizing the continuity of read accesses.

\begin{table}[t]
    \small
    \centering
    \caption{Latency (ms) and bandwidth (MB/s) across different ratios of non-activated neurons in OPT-350M on OnePlus 12. Speedups are calculated relative to the latency of dense case.}
    \label{tab:background-sparsity}
    \begin{tabular}{l|c|c|c|c|c}
        \toprule
        \textbf{Ratio}     & dense   & 10\%    & 20\%    & 30\%   & 40\%   \\
        \midrule
        \textbf{Bandwidth} & 1637.61 & 1355.35 & 1089.24 & 904.69 & 746.03 \\
        \textbf{Latency}   & 234.49  & 254.96  & 281.96  & 297.10 & 308.76 \\
        \textbf{Speedup}   & -       & 0.92    & 0.83    & 0.79   & 0.76   \\
        \midrule
        \textbf{Ratio}     & 50\%    & 60\%    & 70\%    & 80\%   & 90\%   \\
        \midrule
        \textbf{Bandwidth} & 598.82  & 524.50  & 441.33  & 396.43 & 368.05 \\ 
        \textbf{Latency}   & 320.63  & 292.78  & 260.86  & 193.68 & 104.18 \\
        \textbf{Speedup}   & 0.73    & 0.80    & 0.90    & 1.21   & 2.25   \\
        \bottomrule
    \end{tabular}
\end{table}

\subsection{Analysis: IOPS as the Bottleneck}
Activation sparsity enables smartphones to accommodate larger and more powerful models within the limited DRAM capacity required for smaller models by activating only a subset of parameters. However, this memory saving comes at the cost of increased I/O operations between flash memory and DRAM. As listed in Table~\ref{tab:intro-io-bottleneck}, I/O operations account for the majority of total inference latency, severely constraining overall inference efficiency. Consequently, the efficiency of I/O operations emerges as a pivotal determinant of enabling the smooth deployment of LLMs on smartphones.

The root cause of the increased I/O operations lies in the dynamic nature of activation sparsity, where the subset of activated parameters varies with each model input. As a result, every new inference request generates I/O operations to transfer new activated parameters. More critically, mitigating this I/O overhead presents significant challenges. First, identifying the activated parameters in a given layer depends on the outputs of the previous layer, making it difficult to overlap I/O latency with computation. Second, the disjoint activation of parameters across different inference requests diminishes the effectiveness of caching strategies. Although the volume of data transfer can be reduced, the remaining parameters still suffer from low I/O bandwidth utilization.

Table~\ref{tab:background-sparsity} demonstrates that the bottleneck in I/O operations arises primarily from low effective bandwidth utilization, rather than the volume of data transfer. As the ratio of non-activated neurons increases, the I/O volume between flash memory and DRAM decreases. However, inference latency remains longer than that of the fully-activated (dense) case until the non-activation ratio reaches around 80\%. This inefficiency occurs because, in conventional model-structure-based placements, activated parameters are scattered across flash memory, leading to small-grained read accesses that limit bandwidth utilization. Therefore, I/O operations become heavily IOPS-bound, preventing the full exploitation of the available UFS bandwidth on smartphones.

Drawing from these observations, we derive a crucial insight: the conventional placement of model parameters in flash memory, guided by model structure, is misaligned with the dynamic activation sparsity during LLM inference. Consequently, the key to addressing the I/O bottleneck lies in designing an optimized placement strategy for parameters, one that maximizes the continuity of read accesses and enables the full exploitation of available UFS bandwidth.

\begin{figure}[t]
    \centering
    \includegraphics[width=1.0\linewidth]{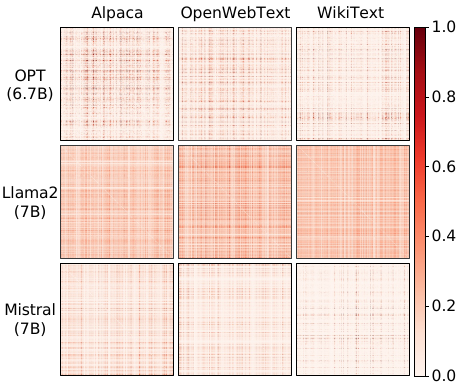}
    \caption{Visualization of neuron co-activation patterns across different LLMs (vertical) and datasets (horizontal). Each matrix represents the adjacency matrix of neurons within a layer of a given LLM, where the element at position $(i, j)$ indicates the co-activation frequency between neuron $i$ and neuron $j$. Brighter colors denote high values.}
    \label{fig:method-co-activation}
\end{figure}

\section{\Neuralink{} Overview}
We propose \Neuralink{}, an efficient approach to accelerating LLM inference on smartphones with optimized I/O accesses. Unlike previous studies that primarily focus on the efficiency of either computation or cache management, \Neuralink{} addresses the I/O bottleneck by directly enhancing the I/O bandwidth between flash memory and DRAM.

\begin{figure}[t]
    \centering
    \includegraphics[width=1.0\linewidth]{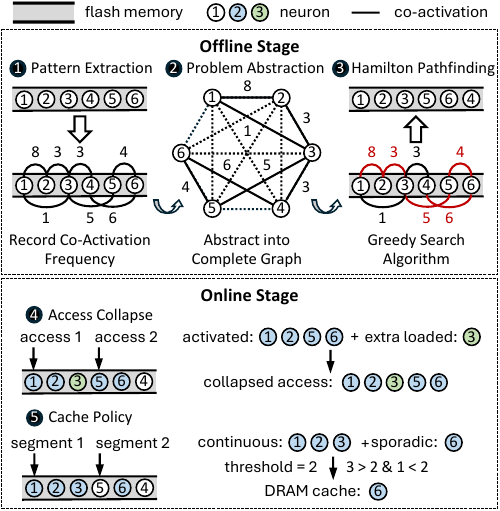}
    \caption{Overview of \Neuralink{}, which operates in two stages: an offline stage and an online stage.}
    \label{fig:method-overview}
\end{figure}

The design of \Neuralink{} is rooted in \textit{Neuron Co-Activation}, a phenomenon prevalent in activation sparsity yet underexplored in current works. As presented in Figure~\ref{fig:method-co-activation}, neurons in LLMs exhibit strongly correlated activation patterns across different model structures and datasets. Although similar observations have been validated in previous studies~\cite{llm-flash, powerinfer-2}, this phenomenon remains largely underexplored due to its intrinsic complexity. By incorporating both algorithmic and system-level optimizations, \Neuralink{} is the first to leverage neuron co-activation for optimizing flash memory placement. Figure~\ref{fig:method-overview} presents an overview of \Neuralink{}.

\noindent\textbf{Offline Correlation-Aware Clustering (\S~\ref{sec:offline}).} \Neuralink{} begins with the identification of an optimized neuron placement in flash memory. The core idea in this stage is to cluster frequently co-activated neurons together, which involves three key steps: \scalebox{1.2}{\ding{182}} \textbf{Pattern Extraction.} We propose a specific design to efficiently extract neuron co-activation patterns based on profiling results. These extracted patterns quantify the strength of co-activation correlations among neurons, providing the foundation for subsequent neuron rearrangement. \scalebox{1.2}{\ding{183}} \textbf{Problem Abstraction.} We model the neuron placement in flash memory as a graph representation. Building on this abstraction, we reformulate the optimization task into a classic Hamiltonian pathfinding problem. This conversion allows us to apply efficient graph-theoretic techniques, leading to a more effective solution for neuron placement optimization. \scalebox{1.2}{\ding{184}} \textbf{Hamiltonian Pathfinding.} Given the NP-complete nature of the optimization problem, we devise a heuristic algorithm that greedily searches for the optimal placement. We prove that our algorithm can find a locally optimal solution with a polynomial time complexity.

\noindent\textbf{Online Continuity-Centric Processing (\S~\ref{sec:online}).} \Neuralink{} further employs custom data access and DRAM management techniques at runtime, complementing the offline design to facilitate more continuous read accesses. \scalebox{1.2}{\ding{185}} \textbf{Access Collapse.} Despite the optimized neuron placement, the inherent dynamics in neuron activation lead to unavoidable discontinuous read access at runtime. To address this, we propose strategically merging nearby discontinuous read accesses by loading additional neurons between them. This merging approach, with minimal overhead, effectively reduces the frequency of discontinuous read access. \scalebox{1.2}{\ding{186}} \textbf{Cache Policy.} Retaining the most frequently activated neurons in DRAM effectively reduces repeated neuron transfers. However, this approach alone risks disrupting the continuity of optimized neuron placement in flash memory. To mitigate this, we propose caching neurons in DRAM at the granularity of neuron segments, rather than individual neurons. This strategy helps prevent fragmentation in flash memory while introducing minimal modification to the existing caching framework.

\begin{figure*}
    \centering
    \includegraphics[width=1.0\linewidth]{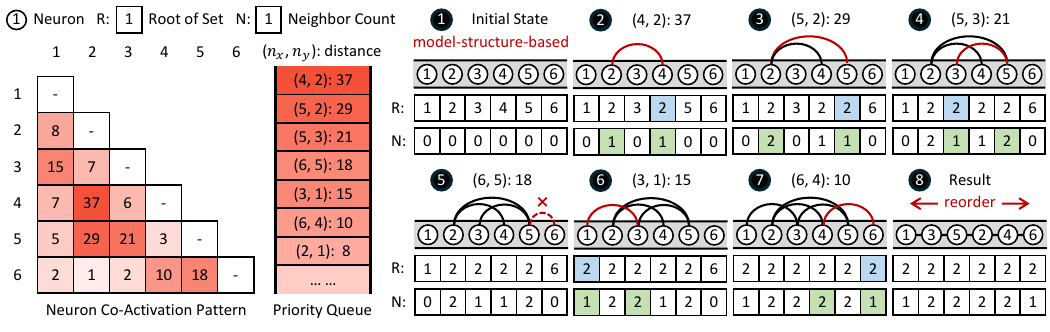}
    \caption{An example of neuron placement search algorithm. Beginning with the neuron co-activation patterns, which are profiled offline (represented as the co-activation frequency in the figure), the algorithm computes the distance between each pair of neurons and ranks them using a priority queue. These pairs are then sequentially retrieved and merged (if possible) at each step. Finally, a single link that contains all neurons is formed, serving as the new neuron placement in flash memory.}
    \label{fig:method-offline}
\end{figure*}

\section{Offline Correlation-Aware Clustering}
\label{sec:offline}
\subsection{Step 1: Parameter-Efficient Pattern Extraction}
LLMs~\cite{opt,gpt3,llama2,mistral,deepseek-v3} are typically based on transformer architectures, consisting of two key components: Multi-Head Attention (MHA) block and Feed Forward Network (FFN) block. In \Neuralink{}, we focus primarily on offloading the parameters of FFN blocks to flash memory, while remaining the MHA block in DRAM. On one hand, recent LLMs are adopting attention variants like Group Query Attention~\cite{gqa}, Multi Query Attention~\cite{mqa}, and Multi-head Latent Attention~\cite{mla} to enhance inference efficiency. These mechanisms significantly reduce the parameter overhead of the attention block. On the other hand, the Mixture-of-Experts (MoE) mechanism becomes increasingly prevalent in LLMs~\cite{deepseekmoe,switch-transformer,mixtral-of-experts}, effectively managing computational costs when scaling up model parameters in FFN blocks. Consequently, the FFN block emerges as the dominant factor in parameter size.

To extract the neuron co-activation patterns, we initially utilize an \textit{Adjacency Matrix} to record the activation frequencies of neurons in LLMs. This step is performed only once, prior to inference, utilizing a dataset associated with the upcoming tasks. By estimating probability using frequency $f$, we compute the activation probability of the neuron $n_{i}$, denoted as $P(i)$, and the co-activation probability of neuron $n_{i}$ and neuron $n_{j}$, denoted as $P(ij)$, as follows:
\begin{equation}
\label{eq:p_neuron}
    P(i) = \frac{f(n_i)}{\sum_{k=1}^{N}{f(n_k)}} 
\end{equation}
\begin{equation}
\label{eq:p_neuron_pair}
    P(ij) = \frac{f(n_i,n_j)}{\sum_{k=1}^{N}\sum_{l=1}^{N}{f(n_k,n_l)}} 
\end{equation}
Here, $N$ denotes the number of neurons in a weight matrix. When performing statistics, \Neuralink{} accounts for the bundling relationship between neurons across different weight matrices in the same FFN block. For instance, in the ReLU-based FFN block~\cite{opt}, the columns of up projection matrix are bundled with the corresponding rows of down projection matrix, as their activations both rely on whether the same intermediate values are zero or not. A similar bundling relationship exists among the gate, up, and down projection matrices in the SwiGLU-based FFN block~\cite{llama2,mistral}.

\subsection{Step 2: Graph-Based Problem Abstraction}
Following the extraction of neuron co-activation patterns, the next step is to determine an optimized neuron placement in flash memory. To enable more continuous read access, neurons that frequently co-activate should ideally be positioned in nearby memory addresses. Given the immense number of neurons in LLMs, the potential neuron placements are extremely large. Innovatively, we reformulate this problem as a graph representation, providing a pathway to a more efficient solution by leveraging graph-theoretic techniques.

\noindent\textbf{Graph Abstraction.} We model the co-activation relationships between neurons using a \textit{Complete Graph}. In this graph, each node represents a neuron, and each edge represents the co-activation relationship between two neurons. Specifically, we define the value of each edge as the \textit{Distance Between Two Neurons}, which reflects the degree of their co-activation:
\begin{equation}
\label{eq:dist-neuron}
    \text{dist}(n_i,n_j) \vcentcolon = 1 - P(ij)
\end{equation}
According to this definition, the more frequently two neurons are co-activated, the smaller their distance.

\noindent\textbf{Hamilton Pathfinding.} The objective of \Neuralink{} is to minimize the expected number of I/O operations for a given inference request. By leveraging the graph-based abstraction, this objective naturally transforms into the problem of \textit{identifying the shortest Hamiltonian path~\cite{hamiltonian-path} in a complete graph}. First, the Hamiltonian path, by definition, visits each node exactly once, thereby ensuring that all neurons stored in flash memory are involved. Second, the shortest Hamiltonian path serves as the optimization criterion, aligning with the objective of maximizing the likelihood of clustering co-activated neurons together. This reformulation enables \Neuralink{} to leverage advanced graph-theoretic techniques for efficient neuron placement optimization.

\subsection{Step 3: Heuristic-Driven Search Algorithm}
In graph theory, determining the shortest Hamiltonian path in a graph is an NP-complete problem~\cite{np-complete}. To address this computational challenge, we design a heuristic algorithm that efficiently searches for an optimized neuron placement with a polynomial time complexity. Figure~\ref{fig:method-offline} illustrates a simplified example of our approach.

\noindent\textbf{Algorithm Details.} The core idea of the algorithm is to treat a neuron placement as a neuron link and iteratively merge them until all neurons are connected within a single link. To minimize the expected number of I/O operations, we employ a greedy merging strategy that prioritizes merging the nearest links at each step. Formally, based on the definition of \textit{Distance Between Two Neurons} in Equation~\ref{eq:dist-neuron}, we define the \textit{Distance Between Two Neuron Links} as follows:
\begin{equation}
\label{eq:dist-link}
\begin{split}
    \text{dist}(l_i, l_j) \vcentcolon = \min \{ \text{dist}(l_i(h), l_j(h)), \text{dist}(l_i(h), l_j(t)), \\ 
                                    \text{dist}(l_i(t), l_j(h)), \text{dist}(l_i(t), l_j(t)) \}
\end{split}
\end{equation}
Here, $l_i(h)$ and $l_i(t)$ denote the head and tail neurons (arbitrary, as the link $l_i$ is undirected) of a neuron link $l_i$, respectively. The equation implies that the distance between two neuron links is determined by the shortest distance between either the head or tail neuron of the two links.

Algorithm~\ref{algo:method-offline} outlines the process in pseudocode. The algorithm begins by taking a set of neurons $\mathcal{N}$ and their co-activation probability $P(ij)$ as input (Line 1). Initially, each neuron in $\mathcal{N}$ is treated as an individual link (Line 2), and the distance between each pair of neurons is computed based on Equation~\ref{eq:dist-neuron} (Lines 7-8). The algorithm then proceeds iteratively, searching for the nearest pair of links to merge in a greedy manner (Lines 9-18). For each pair of links, the distance is computed based on Equation~\ref{eq:dist-link} (Lines 10-12), and the pair with the smallest distance is selected for merging (Lines 13-18). This process repeats until only a single link remains, which contains all the neurons (Line 9). Finally, the merged link is returned as the output of the algorithm, representing the optimized neuron placement $\mathcal{P}$ (Lines 19-24).

\noindent\textbf{Implementation Details.} We leverage two advanced data structures, the priority queue~\cite{priority-queue} and the disjoint set~\cite{disjoint-set}, to enhance the efficiency of the algorithm. The priority queue is used to efficiently identify the nearest neuron link, and the disjoint set union facilitates the efficient management and merging of neuron links. All neurons within the same link share the same root in the disjoint set, which is updated after each set merge. Additionally, we maintain a count of the neighbors for each neuron, $\mathrm{NbrCnt}$, to classify whether a neuron is inside or at the edge of a link ($\mathrm{NbrCnt} = 2 \text{ or } 1$).

\begin{algorithm}[t]
    \small
    \caption{Neuron Placement Search Algorithm}
    \label{algo:method-offline}
    \begin{algorithmic}[1]
        \State \textbf{Input:} Neuron set $\mathcal{N}$, Neuron Co-activation Probability $P(ij)$
        \State \textbf{Output:} Optimized neuron placement $\mathcal{P}$
        \Function{GreedySearch}{$\mathcal{N}$}
            \State Initialize $\textrm{NbrCnt}[n] \gets 0$ for all $n \in \mathcal{N}$
            \State Initialize priority queue $Q \gets \emptyset$
            \State Initialize disjoint sets $S(n)$ for all $n \in \mathcal{N}$
            \For{each pair $(n_i, n_j) \in \mathcal{N} \times \mathcal{N}, n_i \neq n_j$}
                \State $Q.\text{push}((n_i, n_j), \text{dist}(n_i, n_j))$
            \EndFor
            
            \While{$Q \neq \emptyset$}
                \State $(n_x, n_y) \gets Q.\text{pop()}$
                \If{$\textrm{NbrCnt}[n_x] = 2$ \textbf{or} $\textrm{NbrCnt}[n_y] = 2$}
                    \State \textbf{continue}  \Comment{Skip if either neuron is inside a link}
                \EndIf
                \State $\textrm{root}_x \gets \textrm{Find}(n_x)$, $\textrm{root}_y \gets \textsc{Find}(n_y)$
                \If{$\textrm{root}_x \neq \textrm{root}_y$}
                    \State $\textrm{NbrCnt}[n_x]$++
                    \State $\textrm{NbrCnt}[n_y]$++
                    \State \textrm{Union}($\textrm{root}_x$, $\textrm{root}_y$)
                    \State $\textrm{Link}(n_x,n_y)$  \Comment{Update neuron linkings}
                \EndIf
            \EndWhile

            \State $\mathcal{P} \gets []$
            \State $c \gets \text{Select first neuron from } \{n \in \mathcal{N} \mid \text{NbrCnt}[n] = 1\}$ 
            \LeftComment{Set current neuron to starting point}
            
            \While{$c \neq s$ \textbf{and} $\rm NbrCnt[c] \neq 1$}
                \State $\mathcal{P}.\text{append}(c)$  \Comment{Add $c$ to the optimized placement}
                \State $c \gets \textrm{NextNeuron}(c)$  \Comment{Move to next neuron linked to $c$}
            \EndWhile

            \State \Return $\mathcal{P}$
        \EndFunction
    \end{algorithmic}
\end{algorithm}

\noindent\textbf{Complexity Analysis.} The primary computational complexity of the algorithm arises from the $n^2$ pairwise enumeration of $n$ neurons. For each pair of neurons, both the insertion and pull operations in the priority queue have a time complexity of $O(\log n)$. Meanwhile, both the operations for finding a root and merging two sets in the disjoint set have a time complexity of $O(1)$. Consequently, the overall time complexity of the algorithm is $O(n^2 \log n)$.

\section{Online Continuity-Centric Processing}
\label{sec:online}
Through offline correlation-aware clustering, neurons that are frequently co-activated are strategically placed contiguously in flash memory. However, the dynamic and intricate nature of neuron co-activation makes static neuron placement alone insufficient to entirely alleviate IOPS limitations. To fully exploit the flash memory and DRAM resources to serve neuron read requests, we design specific online processing techniques that focus on maintaining continuity of access. These techniques are aimed at addressing two primary challenges, manifesting in data access and caching.

The first challenge arises from inherent activation dynamics. Due to the stochastic nature of neuron activation, it is infeasible to consistently follow the co-activation patterns extracted offline. Although neurons that are frequently co-activated are placed in close positions, minor variances induced by these dynamics can still lead to unavoidable discontinuous read access during runtime. Particularly, some read access may occur near each other but be split into two I/O operations by just a few activated neurons in between.

The second challenge stems from misaligned cache strategies. Conventional cache strategies typically treat neurons individually, caching only the most frequently activated neurons without considering their placement in flash memory. This oversight potentially breaks the continuity of neuron placement in flash memory, undermining our optimizations for I/O operation reduction. Moreover, directly caching all co-activated neurons in a continuous manner could consume excessive DRAM resources, reducing cache efficiency.

\subsection{IOPS-Friendly Access Collapse}
\begin{figure}[t]
    \centering
    \includegraphics[width=1.0\linewidth]{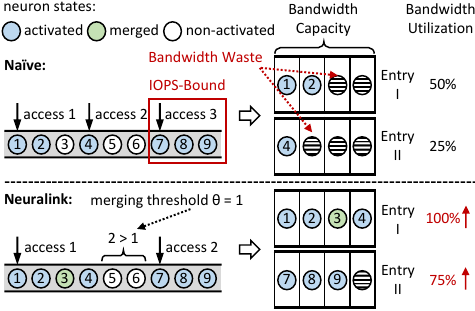}
    \caption{Static neuron placement suffers from IOPS-bound due to dynamic activation. \Neuralink{} mitigates this by strategically merging intervening neurons between accesses.}
    \label{fig:method-online}
\end{figure}

In \Neuralink{}, we introduce an innovative online technique that strategically combines nearby read accesses. The fundamental insight driving this approach is that while co-activated neurons cannot always be placed contiguously, they are likely to be positioned in close proximity following offline correlation-aware clustering. As illustrated in Figure~\ref{fig:method-online}, consider a scenario where neurons $n_1$, $n_2$, $n_3$, and $n_4$ are stored contiguously, but occasionally only $n_1$, $n_2$, and $n_4$ are activated, necessitating two distinct read operations. However, when IOPS is limited, the inclusion of more neurons per read operation yields superior overall performance. Capitalizing on this observation, when two disjoint but proximate neuron links are co-activated, we speculatively read the intervening neurons. This strategy effectively coalesces the two separate neuron links into a single, contiguous read access, thereby substantially enhancing overall efficiency.

The execution of this IOPS-friendly access collapse is governed by two key factors during runtime. (1) \textbf{Bandwidth Trade-off.} Introducing additional neurons for merging involves a trade-off between increasing the data transfer size and reducing IO operations, with the goal of enhancing bandwidth utilization. We employ a threshold-based approach: if the number of neurons between two neuron links falls below a predefined threshold, collapse is performed; otherwise, it is skipped. This approach ensures that the inclusion of additional neurons occurs only when it provides a benefit for bandwidth enhancement, without causing bandwidth waste. (2) \textbf{Storage Bottleneck.} While access collapse can reduce IO operations, it only improves bandwidth efficiency if the storage is IOPS-bound rather than bandwidth-bound. To handle this, we implement a bottleneck detector that periodically checks whether the achieved bandwidth has reached the maximum capacity of storage. If the bandwidth is fully utilized, the system reverts to the original access strategy.

\subsection{Linking-Aligned Cache Policy}
It is natural to store the neurons that are most frequently activated in DRAM to reduce the redundant data transfer between flash memory and DRAM. However, directly applying existing cache policies is inefficient in \Neuralink{}, as these policies typically operate at the level of individual neurons, disregarding both neuron co-activation patterns and the optimized placement in flash memory. For example, suppose neurons $n_1$, $n_2$, $n_3$, and $n_4$ are stored together and often co-activate. If $n_2$ is more frequently activated than the others, it will have a higher probability of being cached, leading to discontinuous fragments in flash memory. A potential solution is to cache the neurons that are stored continuously in flash memory together, reducing the likelihood of this situation. However, this approach would occupy a large portion of DRAM, making it impractical for smartphones.

To address this, we introduce an additional layer of access management to the existing cache policies, requiring minimal modification while improving overall efficiency. In \Neuralink{}, activated neurons are divided into two categories: outlier neurons and continuous segments. Outlier neurons, as the name suggests, are those neurons being co-activated with only a few surrounding neurons. In contrast, continuous segments consist of a series of neurons that are activated together in succession. Considering both resource usage and efficiency enhancement, \Neuralink{} devises different strategies for the two categories of neurons. For outlier neurons, \Neuralink{} caches them as usual. However, continuous segments are cached with a lower probability compared to outlier neurons. This is primarily because caching continuous segments requires more memory resources and provides less efficiency enhancement (already benefit from continuous access). Furthermore, if some neurons in these segments are evicted while others remain in the cache, it will lead to discontinuous reads in flash memory. Although the waste of IOPS can be alleviated by access collapse, this still results in wasted DRAM resources. Our cache policy works seamlessly with state-of-the-art cache designs, as we only control the caching admitting policy while leaving the rest unchanged.

\section{Evaluation}
\subsection{Experimental Setup}
\noindent\textbf{Hardware.} We conduct evaluations across a diverse set of smartphones, as detailed in Table~\ref{tab:exp-configuration-hardware}. These devices feature varying SoC configurations, memory capacities, and UFS versions, spanning from low-tier to high-tier hardware. All experiments are conducted on Termux~\cite{termux}, an Android shell application, to ensure compatibility across platforms.

\noindent\textbf{Models.} We choose seven widely adopted LLMs for evaluation, as outlined in Table~\ref{tab:exp-configuration-model}. These models vary in architecture and size, with each utilizing activation sparsity through well-developed techniques~\cite{deja-vu,pro-sparse,turbo-sparse,cats,slm-activation-sparsity}. Besides LLMs using ReLU-family activation functions, we also evaluate SiLU-family LLM (MobiLlama) and GeLU-family LLM (Phi-2). All models in our experiments are evaluated in FP16 precision, consistent with the practices in recent research~\cite{llm-flash,powerinfer}.

\noindent\textbf{Datasets.} We evaluate \Neuralink{} using a wide range of datasets, covering categories such as plain text, instructions, math or commonsense tasks. These datasets encompass diverse linguistic structures, enabling a comprehensive evaluation across varying sparsity patterns. For each dataset, we randomly select 1,024 samples from the training set for profiling and perform evaluation on the test set.

\begin{table}[t]
    \small
    \centering
    \caption{Smartphone hardware configurations.}
    \label{tab:exp-configuration-hardware}
    \begin{adjustbox}{width=1.0\linewidth,center}
    \setlength{\tabcolsep}{3pt}
    \begin{tabular}{lllll}
        \toprule
        \textbf{Device} & \textbf{SoC}        & \textbf{DRAM} & \textbf{Flash} & \textbf{Storage} \\ \midrule
        OnePlus 12      & Snapdragon 8 Gen 3  & 24GB          & 1TB            & UFS4.0           \\
        OnePlus Ace 3   & Snapdragon 8 Gen 2  & 16GB          & 512GB          & UFS4.0           \\
        OnePlus Ace 2   & Snapdragon 8+ Gen 1 & 16GB          & 512GB          & UFS3.1           \\
        \bottomrule
    \end{tabular}
    \end{adjustbox}
\end{table}

\begin{table}[t]
    \small
    \centering   
    \caption{Model configurations.}
    \label{tab:exp-configuration-model}
    \begin{adjustbox}{width=0.98\linewidth,center}
    \begin{threeparttable}
    \begin{tabular}{llllr}
        \toprule
        \textbf{Model}                  & \textbf{Size} & \textbf{Neuron}\tnote{\dag} & \textbf{ActFn} & \textbf{Sparsity Method}                         \\ \midrule
        OPT-S~\cite{opt}                & 350M          & 8,192                       & ReLU           & Prediction~\cite{deja-vu}                        \\
        OPT-M~\cite{opt}                & 1.3B          & 16,384                      & ReLU           & Prediction~\cite{deja-vu}                        \\
        OPT-L~\cite{opt}                & 6.7B          & 32,768                      & ReLU           & Prediction~\cite{deja-vu}                        \\
        Llama-2~\cite{llama2}           & 6.5B          & 33,024                      & fatReLU        & Prediction~\cite{pro-sparse}                     \\
        Mistral~\cite{mistral}          & 7.8B          & 43,008                      & dReLU          & Prediction~\cite{turbo-sparse}                   \\
        MobiLlama~\cite{mobillama}      & 1.0B          & 16,896                      & SiLU           & Threshold~\cite{cats}                            \\
        Phi-2~\cite{phi-2}              & 2.7B          & 20,480                      & GeLU           & Threshold~\cite{slm-activation-sparsity}         \\
        \bottomrule
    \end{tabular}
    \begin{tablenotes}
    \item [\dag]The number of neurons per FFN block.
    \end{tablenotes}
    \end{threeparttable}
    \end{adjustbox}
\end{table}

\begin{table}[t]
    \small
    \centering   
    \caption{Dataset configurations.}
    \label{tab:exp-configuration-dataset}
    \begin{adjustbox}{width=1.0\linewidth,center}
    \setlength{\tabcolsep}{2pt}
    \begin{tabular}{ll}
        \toprule
        \textbf{Dataset}               & \textbf{Description} \\ \midrule
        Alpaca~\cite{alpaca}           & Instructions-tuning dataset for Llama \\
        OpenWebText~\cite{openwebtext} & High-quality web-scraped text corpus \\
        WikiText~\cite{wikitext}       & Wikipedia-based language modeling dataset \\
        GSM8K~\cite{gsm8k}             & Grade-school math problem dataset \\
        HellaSwag~\cite{hellaswag}     & Commonsense reasoning benchmark \\
        MMLU~\cite{mmlu}               & Multitask knowledge evaluation benchmark \\
        SWAG~\cite{swag}               & Commonsense reasoning sentence completion \\
        PIQA~\cite{piqa}               & Physical commonsense reasoning benchmark \\
        \bottomrule
    \end{tabular}
    \end{adjustbox}
\end{table}

\noindent\textbf{Baselines.} We benchmark \Neuralink{} with two state-of-the-art LLM inference frameworks on smartphones. The first, llama.cpp~\cite{llama-cpp}, is the most widely used CPU framework with state-of-the-art performance. The second baseline, LLMFlash~\cite{llm-flash}, is specially designed for the activation sparsity paradigm. Since LLMFlash is not open-source, we port it into llama.cpp by adding support for controlled parameter offloading and integrating its key I/O optimizations, such as row-column bundling. Additionally, we introduce a simplified variant of \Neuralink{} (\textsc{Neuralink-S}), which directly arranges neuron placement in decreasing order of activation frequency. This comparison further underscores the effectiveness of correlation-aware clustering in \Neuralink{}.

\noindent\textbf{Metrics.} We use the end-to-end latency per token as our primary performance metric and measure I/O bandwidth for a finer analysis. Specifically, we use the kernel io\_uring~\cite{liburing} to load data from flash memory, measuring I/O latency as the time from the first request issue to the last, and calculating bandwidth based on the total data read. Notably, bandwidth here refers to the effective bandwidth, which only considers the activated neurons. For all LLMs, we offload their FFN blocks to flash memory and maintain a 0.05 cache ratio in DRAM. We normalize the metrics values when large discrepancies occur for clarity. All metrics are averaged over 128 token generations, repeated across 10 trials.

\subsection{Overall Performance}
\begin{figure*}[t]
    \centering
    \includegraphics[width=1.0\linewidth]{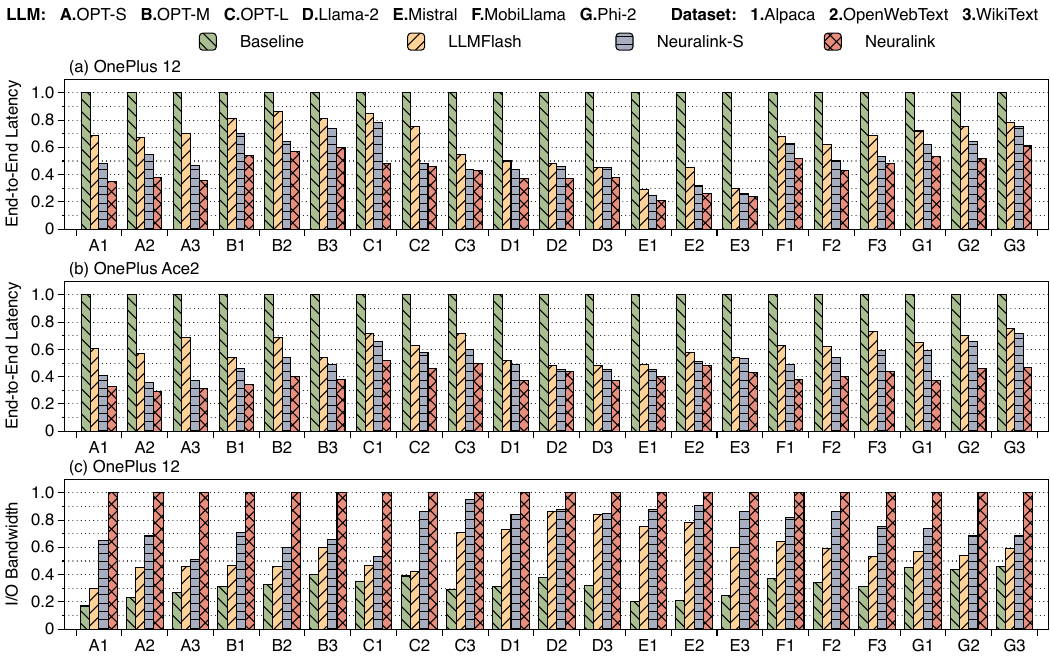}
    \caption{Overall performance (normalized) across various LLMs and datasets. (a) and (b): End-to-end latency on OnePlus 12 and Ace2; (c): I/O bandwidth on OnePlus 12. Each data point ($A1$) on x-axis denotes a combination of LLM ($A$) and dataset ($1$).}
    \label{fig:exp-overall-performance}
\end{figure*}

\noindent\textbf{End-to-end Latency.} Figures~\ref{fig:exp-overall-performance}\hyperref[fig:exp-overall-performance]{a} and \ref{fig:exp-overall-performance}\hyperref[fig:exp-overall-performance]{b} present the end-to-end performance speedups across different LLMs and datasets on OnePlus 12 and OnePlus Ace2. Besides, Table~\ref{tab:exp-sparsity-level} provides the corresponding sparsity levels (i.e., the ratio of activated neurons to the total) across all experiments. For prediction-based methods, sparsity levels are dynamically determined at runtime by well-trained predictors~\cite{deja-vu,pro-sparse,turbo-sparse}. For sparsity methods like threshold and attribution score, the sparsity levels are pre-defined. Following prior research~\cite{cats,slm-activation-sparsity}, we set these levels as 30\% and 20\%, respectively. The results demonstrate that \Neuralink{} consistently outperforms the baselines, achieving average speedups of $2.37\times$, $1.48\times$, and $1.25\times$ over llama.cpp, LLMFlash, and \textsc{Neuralink-S}, respectively. Moreover, the results confirm that activation sparsity is prevalent across various LLM and dataset combinations, with sparsity levels varying. \Neuralink{} achieves greater performance improvements at higher sparsity levels, such as in OPT models. This is because reduced neuron activation exacerbates scattered flash memory access, making efficient I/O operations more critical. As LLMs become denser, they more easily reach the bandwidth limitations of the device. However, the bandwidth underutilization persists and \Neuralink{} still achieves an average $1.39\times$ speedup over LLMFlash for non-OPT models on two different hardware. 

\begin{table}[t]
    \small
    \centering   
    \caption{The ratios of activated neurons to the total neurons during inference across various LLMs and datasets.}
    \label{tab:exp-sparsity-level}
    \begin{tabular}{lccc}
        \toprule
        \textbf{Model} & \textbf{Alpaca} & \textbf{OpenWebText} & \textbf{WikiText} \\ \midrule
        OPT-S          &  9.48\%         &  9.40\%              &  9.73\%           \\
        OPT-M          &  3.65\%         &  3.88\%              &  3.90\%           \\
        OPT-L          &  3.17\%         &  3.30\%              &  2.97\%           \\
        Llama-2        & 23.67\%         & 27.64\%              & 21.27\%           \\
        Mistral        & 23.54\%         & 23.99\%              & 24.51\%           \\
        MobiLlama      & 30.00\%         & 30.00\%              & 30.00\%           \\
        Phi-2          & 20.00\%         & 20.00\%              & 20.00\%           \\
        \bottomrule
    \end{tabular}
\end{table}

\noindent\textbf{Achieved Bandwidth.} Figure~\ref{fig:exp-overall-performance}\hyperref[fig:exp-overall-performance]{c} presents the detailed I/O bandwidth results on OnePlus 12. The results show that \Neuralink{} achieves average bandwidth improvements of $3.28\times$, $1.80\times$, and $1.36\times$ over the three baselines. These findings demonstrate that \Neuralink{} effectively shifts the I/O bottleneck from IOPS to bandwidth across various sparsity levels. Furthermore, the strong conversion efficiency of I/O optimizations into end-to-end latency not only confirms that I/O operations are the primary bottleneck but also highlights that \Neuralink{} introduces minimal runtime overhead.

\subsection{Performance Breakdown}
\begin{figure}[t]
    \centering
    \includegraphics[width=1.0\linewidth]{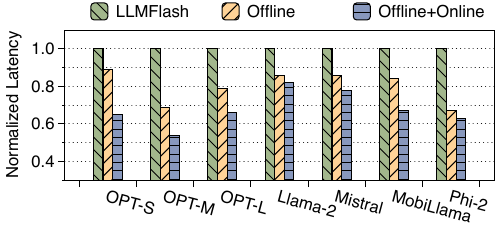}
    \caption{Performance breakdown of \Neuralink{}. The results are evaluated using Alpaca dataset on OnePlus Ace2.}
    \label{fig:exp-performance-breakdown}
\end{figure}

As shown in Figure~\ref{fig:exp-performance-breakdown}, we evaluate the effectiveness of two key components of \Neuralink{}. Starting with the strong baseline LLMFlash, we measure the end-to-end latency across various LLMs. The results show that the offline and online components provide average speedups of $1.32\times$ and $1.20\times$, respectively. These findings align with the design of \Neuralink{}, where the offline component serves as the cornerstone and the online component plays a supplementary role.

\begin{table}[t]
    \small
    \centering   
    \caption{Offline execution time (in seconds) of the neuron placement search algorithm across various LLMs and datasets using an AMD EPYX 7V13 processor.}
    \label{tab:exp-ablation-offline-overhead}
    \begin{tabular}{lccc}
        \toprule
        \textbf{Model} & \textbf{Alpaca} & \textbf{OpenWebText} & \textbf{WikiText} \\ \midrule
        OPT-S          & 4.48            & 4.42                 & 4.45              \\
        OPT-M          & 21.54           & 22.14                & 22.21             \\
        OPT-L          & 104.23          & 103.56               & 105.90            \\
        Llama-2        & 50.09           & 51.43                & 55.43             \\
        Mistral        & 102.17          & 90.43                & 93.71             \\
        MobiLlama      & 12.20           & 10.22                & 11.41             \\
        Phi-2          & 49.05           & 45.16                & 46.99             \\
        \bottomrule
    \end{tabular}
\end{table}

\begin{figure*}[t]
    \centering
    \includegraphics[width=1.0\linewidth]{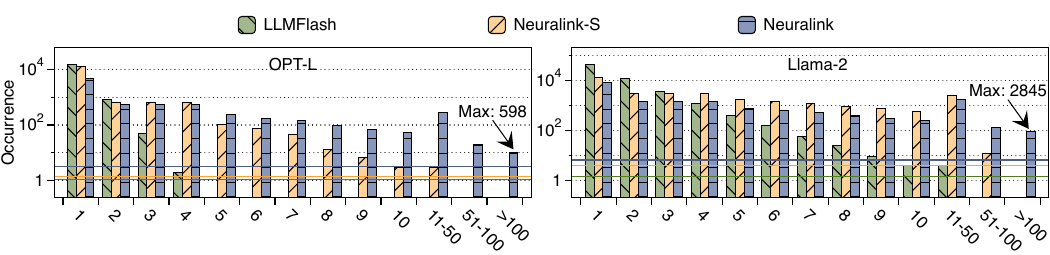}
    \caption{Statistical information on read access lengths per token (unit: neuron bundle) in LLMFlash, \textsc{Neuralink-S}, and \Neuralink{} using Alpaca dataset. A larger value of read access length indicates better continuity. Lines indicate average values.}
    \label{fig:exp-ablation-offline-continuity}
\end{figure*}

\subsection{Offline Ablation Study}
\noindent\textbf{Continuous Access.} The core insight of \Neuralink{} lies in increasing continuous flash memory access by optimizing neuron placement, thereby mitigating I/O overheads. We examine the read access lengths in LLMFlash, \textsc{Neuralink-S}, and \Neuralink{}, as shown in Figure~\ref{fig:exp-ablation-offline-continuity}. LLMFlash, which does not account for neuron co-activation, maintains the read access lengths to below 5 neuron bundles, with average values of 1.06 in OPT-L and 1.42 in Llama-2. \textsc{Neuralink-S} improves continuous access compared to LLMFlash, increasing the averages to 1.34 and 4.03, respectively. However, it considers only individual activation frequency but overlooks the spatial continuity of activated neurons. For example, a neuron may frequently activate but co-activate with different neurons each time. In contrast, \Neuralink{} increases the average read access lengths to 3.12 and 6.62 across two LLMs. Notably, the maximum continuous read access length reaches 589 bundles in OPT-L and 2,845 bundles in Llama-2.

\noindent\textbf{Overhead Analysis.} Table~\ref{tab:exp-ablation-offline-overhead} lists the execution time of the offline search algorithm in \Neuralink{} for optimized neuron placement. We perform experiments across various LLMs and datasets, considering variations in the number of activated neurons. To speed up search process, we implement parallel computation by exploiting the independence of different model layers. The results show that all search processes are complete within a few minutes, even for the largest 7B model. With a polynomial time complexity is $O(n^2\log n)$, the increase in time cost remains modest. Since the search process is performed offline, this overhead is acceptable.

\begin{figure}[t]
    \centering
    \includegraphics[width=1.0\linewidth]{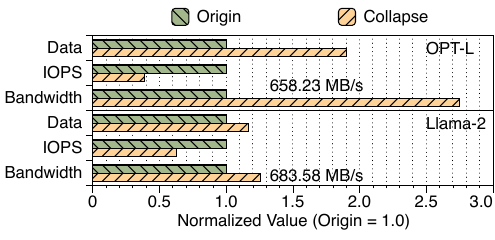}
    \caption{Data transfer volume, IOPS, and bandwidth of activated neurons before and after applying access merging. The experiments use the Alpaca dataset on OnePlus Ace2.}
    \label{fig:exp-ablation-online-merge}
\end{figure}

\subsection{Online Ablation Study}
\noindent\textbf{Access Collapse.} We first evaluate the effectiveness of access collapse. Figure~\ref{fig:exp-ablation-online-merge} shows some I/O information during LLM inference before and after applying access collapse. The results reveal that access collapse provides a bandwidth improvement of $2.75\times$ for OPT-L and $1.26\times$ for Llama-2, with both approaching the peak bandwidth of the device. This improvement comes from a careful trade-off between data transfer volume and IOPS. While additional inactivated neurons are loaded, the reduction in IOPS effectively alleviates the bottleneck and enhances bandwidth utilization.

\begin{figure}[t]
    \centering
    \includegraphics[width=1.0\linewidth]{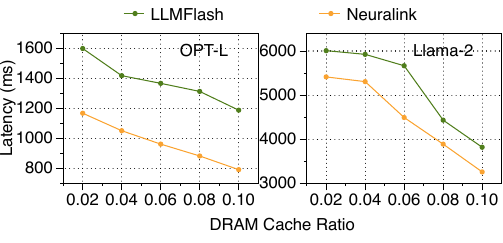}
    \caption{End-to-end latency under varying DRAM cache ratios, evaluated using the Alpaca dataset on OnePlus 12.}
    \label{fig:exp-ablation-online-cache}
\end{figure}

\noindent\textbf{Cache Policy.} We next compare the inference latency of \Neuralink{} with LLMFlash under different cache ratios, as shown in Figure~\ref{fig:exp-ablation-online-cache}. Both LLMFlash and \Neuralink{} utilize the same cache system, S3-FIFO~\cite{s3-fifo}. The results indicate that \Neuralink{} integrates well with existing caching systems, demonstrating a consistent decrease in latency as the cache ratio increases. Besides, \Neuralink{} maintains high cache space efficiency, consistently requiring less DRAM cache space than LLMFLash for the same inference latency. This effectively mitigates the DRAM limitations on smartphones.

\subsection{Sensitivity Analysis}
\begin{figure}[t]
    \centering
    \includegraphics[width=1.0\linewidth]{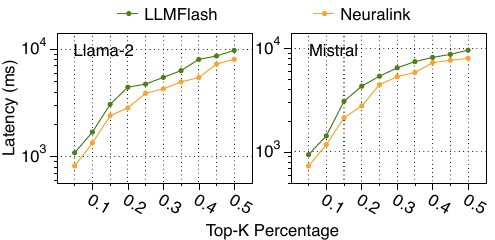}
    \caption{End-to-end latency under different top-K percentages using the Alpaca dataset on OnePlus 12. A higher top-K percentage indicates a lower sparsity level.}
    \label{fig:exp-sensitivity-sparsity}
\end{figure}

\noindent\textbf{Impact of Sparsity Levels.} We first evaluate how variations in sparsity level affect the performance of \Neuralink{}. To control the sparsity level, we apply the top-K sparsity technique~\cite{q-sparse} to the Llama-2 and Mistral, without using their ReLU substitution. Figure~\ref{fig:exp-sensitivity-sparsity} demonstrates that \Neuralink{} achieves greater performance improvements at higher sparsity levels. This is because, when the number of activated neurons is small, their placement in flash memory tends to be more scattered. However, \Neuralink{} continues to reduce end-to-end latency until the sparsity level reaches 50.0\%.

\begin{figure}[t]
    \centering
    \includegraphics[width=1.0\linewidth]{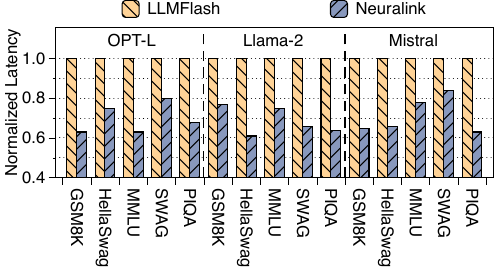}
    \caption{End-to-end latency across more diverse task datasets on OnePlus 12. Same answer lengths are controlled.}
    \label{fig:exp-sensitivity-datasets}
\end{figure}

\noindent\textbf{Impact of Datasets.} We next evaluate the performance of \Neuralink{} across a variety of task datasets, as shown in Figure~\ref{fig:exp-sensitivity-datasets}. The results show that the performance of \Neuralink{} is not sensitive to the choice of datasets. Combined with the findings in Figure~\ref{fig:exp-overall-performance}, these results further confirm that neuron co-activation is widespread in LLM inference and effectively leveraged to optimize I/O operations by \Neuralink{}.

\noindent\textbf{Impact of Profiling.} Ideally, \Neuralink{} should extract neuron co-activation information offline from a profiling dataset that matches the data distribution of the preprocessed dataset. However, in some cases, LLMs may encounter diverse inputs from different tasks during inference. Therefore, we also access the sensitivity of \Neuralink{} to mismatches between the profiling dataset and the testing dataset. As shown in Table~\ref{tab:exp-sensitivity-profiling}, we measure the performance of \Neuralink{} across nine combinations of the profiling and testing datasets, using three different datasets. For example, the first row indicates that \Neuralink{} uses Alpaca as the profiling dataset and is evaluated on Alpaca, OpenWebText, and WikiText during inference. Besides, we report the corresponding speedups of \Neuralink{} compared to LLMFlash (no profiling datasets) across different testing datasets. The results reveal that \Neuralink{} remains effective even when the inputs are biased. Compared to scenarios when the same dataset is used for both profiling and testing, only slight decreases in performance are observed - sometimes even showing improvements. This suggests that neuron co-activation patterns are an intrinsic property of the LLM, with input variations having limited influence. These findings are consistent with recent research on LLM sparsity, which shows that neurons with similar functions tend to activate together~\cite{modularity,sparsing-law}.

\begin{figure}[t]
    \centering
    \includegraphics[width=1.0\linewidth]{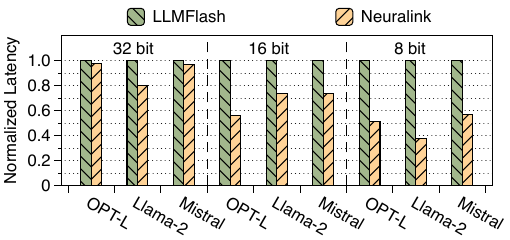}
    \caption{Performance of \Neuralink{} across different parameter precisions using the Alpaca dataset on OnePlus 12.}
    \label{fig:exp-sensitivity-precision}
\end{figure}

\begin{table}[t]
    \small
    \centering   
    \caption{Performance across different combinations of \textit{profiling} (rows) and \textit{testing} (columns) datasets on OnePlus 12. The reported speedup refers to the performance improvement of \Neuralink{} compared to LLMFlash on the testing dataset.}
    \label{tab:exp-sensitivity-profiling}
    \begin{adjustbox}{width=1.0\linewidth,center}
    \setlength{\tabcolsep}{2pt}
    \begin{tabular}{llccc}
        \toprule
        \multirow{2}{*}{\textbf{Model}} & \multirow{2}{*}{\textbf{\textit{Profiling} Dataset}} & \multicolumn{3}{c}{\textbf{\textit{Testing} Dataset}} \\ \cmidrule{3-5}
                                         &             & Alpaca       & OpenWebText  & WikiText     \\ \midrule
        \multirow{6}{*}{OPT-L}           & Alpaca      & 711.09 ms    &  809.80 ms   & 800.46 ms    \\
                                         & Speedup     & $1.86\times$ & $1.65\times$ & $1.59\times$ \\ \cmidrule{3-5}
                                         & OpenWebText & 856.48 ms    &  802.01 ms   & 800.26 ms    \\
                                         & Speedup     & $1.54\times$ & $1.67\times$ & $1.59\times$ \\ \cmidrule{3-5}
                                         & WikiText    & 823.93 ms    & 1031.41 ms   &  784.19 ms   \\
                                         & Speedup     & $1.60\times$ & $1.30\times$ & $1.63\times$ \\
        \midrule
        \multirow{6}{*}{Llama-2}         & Alpaca      & 4405.76 ms   & 4537.75 ms   & 3747.40 ms   \\
                                         & Speedup     & $1.27\times$ & $1.17\times$ & $1.45\times$ \\ \cmidrule{3-5}
                                         & OpenWebText & 3546.12 ms   & 4118.59 ms   & 4318.28 ms   \\
                                         & Speedup     & $1.57\times$ & $1.29\times$ & $1.26\times$ \\ \cmidrule{3-5}
                                         & WikiText    & 4769.36 ms   & 4535.48 ms   & 4578.87 ms   \\
                                         & Speedup     & $1.17\times$ & $1.17\times$ & $1.18\times$ \\
        \bottomrule
    \end{tabular}
    \end{adjustbox}
\end{table}

\noindent\textbf{Impact of Precision.} We evaluate the performance of \Neuralink{} across varying precisions for LLMs. The results show that \Neuralink{} scales efficiently with different precisions, maintaining consistent performance across three LLMs. Notably, \Neuralink{} performs better when LLMs are quantized to lower bit precision, achieving an average speedup of $1.39\times$ when reducing precision from 16-bit to 8-bit. This is because smaller neuron sizes exacerbate the impact of scattered read access, making it harder to reach bandwidth limitations. These findings also highlight the importance of \Neuralink{} in the trend of lower bits utilization on smartphones.

\begin{figure}[t]
    \centering
    \includegraphics[width=1.0\linewidth]{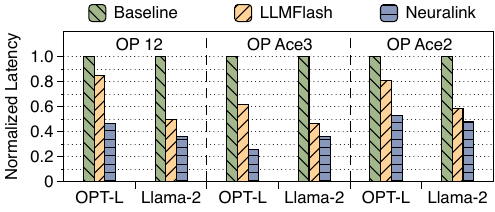}
    \caption{End-to-end latency across more diverse hardware configurations using the Alpaca dataset.}
    \label{fig:exp-sensitivity-hardware}
\end{figure}

\noindent\textbf{Impact of Hardware.} We conclude by evaluating the impact of hardware on \Neuralink{}. Figure~\ref{fig:exp-sensitivity-hardware} shows that \Neuralink{} consistently outperforms two baselines across various hardware configurations. The best performance improvements are observed on OnePlus 12 (OP 12), as its USF storage supports higher bandwidth. Compared to OP 12, OnePlus Ace3 (OP Ace3) shares the same UFS storage but features a less powerful SoC. However, the comparable latency results between OP 12 and OP Ace3 suggest that storage plays a more critical role than the SoC in on-device inference. In contrast, OnePlus Ace2 (OP Ace2) has weaker storage compared to OP 12, making it easier to reach bandwidth limitations.

\section{Related Works}
\noindent\textbf{Model Pruning.} Pruning methods~\cite{pruning-survey-1,pruning-survey-2} aim to reduce the number of model parameters while minimizing performance degradation. Several works~\cite{pruning-static-group,pruning-static-filter,pruning-static-sparse,pruning-static-element,pruning-static-deep-compression} have explored static pruning, where parameters are pruned offline. In contrast, dynamic sparsity methods~\cite{pruning-dynamic-rl,pruning-dynamic-sft,pruning-dynamic-rethink,pruning-dynamic-surgery,pruning-dynamic-low-rank,pruning-dynamic-movement,lemo} determines which parameters to prune at runtime, enabling seamless integration with training or inference. Unlike these methods, \Neuralink{} exploits activation sparsity within LLMs. By retaining all parameters and selectively activating only a subset, it preserves the models' generalization.

\noindent\textbf{Model Quantization.} While aiming for a similar goal, model quantization~\cite{quantization-survey-1,quantization-survey-2} reduces the precision of model parameters by optimizing the utilization of available bits to encode model information more efficiently. Numerous studies have driven precision progressively lower, with efforts ranging from 8-bit~\cite{llm-int8,smooth-quant} to 4-bit~\cite{gptq,zero-quant,bitsandbytes}, 2-bit~\cite{quip}, and even 1-bit~\cite{bitnet,onebit}. However, as precision decreases, data access patterns become increasingly fine-grained and scattered. This, in turn, exacerbates the I/O bottleneck on smartphones.

\noindent\textbf{Sparse Computation Optimization.} Sparse computation typically falls short in performance compared to its dense counterparts. Many works have been proposed to address this. Several compiler-based techniques~\cite{sparta,sparsert} are tailored for static sparsity patterns, while others~\cite{sputnik,cusparse,pit,flash-llm,maxembed} provide support for more general sparsity patterns. In parallel, an increasing number of hardware solutions~\cite{hardware-1,hardware-2,hardware-3,hardware-4,cje,tensor-core} have been specially designed. However, these advancements make the I/O bottleneck more pronounced.

\noindent\textbf{Activation Sparsity Application.} Deja Vu~\cite{deja-vu} pioneered a predictor-based approach for sparsity-based LLM inference. Building upon this, Powerinfer~\cite{powerinfer} exploits this property to enable LLM execution on consumer-grade GPUs by offloading model weights to CPU. In mobile scenarios, LLM in a Flash~\cite{llm-flash} first proposes using flash on smartphones for model offloading. Powerinfer-2~\cite{powerinfer-2} extends this approach further, serving a 47B LLM on a smartphone. However, these methods primarily focus on optimizing DRAM management and overlapping computation with data transfers, achieving only limited bandwidth improvements. \Neuralink{} complements these efforts by enhancing the neuron transfer bandwidth.

\section{Conclusion}
We propose \Neuralink{}, an algorithm-system co-design approach to accelerating LLM inference on smartphones by optimizing neuron placement within flash memory.

\section{Acknowledgement}
We sincerely thank our shepherd Gagandeep Singh and anonymous reviewers for their insightful feedback. The work is supported in part by the National Natural Science Foundation of China under Grant No. 62432004, and by a grant from the Guoqiang Institute, Tsinghua University.

\bibliographystyle{plain}
\bibliography{references}

\end{document}